# EffCNet: An Efficient CondenseNet for Image Classification on NXP BlueBox

**Priyank Kalgaonkar, Mohamed El-Sharkawy**[*]

Department of Electrical and Computer Engineering, Purdue School of Engineering and Technology, Indianapolis, USA

**Email address:**
pkalgaon@purdue.com (P. Kalgaonkar), melshark@purdue.com (M. El-Sharkawy)

[*]Corresponding author

**Abstract:** Intelligent edge devices with built-in processors vary widely in terms of capability and physical form to perform advanced Computer Vision (CV) tasks such as image classification and object detection, for example. With constant advances in the field of autonomous cars and UAVs, embedded systems and mobile devices, there has been an ever-growing demand for extremely efficient Artificial Neural Networks (ANN) for real-time inference on these smart edge devices with constrained computational resources. With unreliable network connections in remote regions and an added complexity of data transmission, it is of an utmost importance to capture and process data locally instead of sending the data to cloud servers for remote processing. Edge devices on the other hand, offer limited processing power due to their inexpensive hardware, and limited cooling and computational resources. In this paper, we propose a novel deep convolutional neural network architecture called EffCNet which is an improved and an efficient version of CondenseNet Convolutional Neural Network (CNN) for edge devices utilizing self-querying data augmentation and depthwise separable convolutional strategies to improve real-time inference performance as well as reduce the final trained model size, trainable parameters, and Floating-Point Operations (FLOPs) of EffCNet CNN. Furthermore, extensive supervised image classification analyses are conducted on two benchmarking datasets: CIFAR-10 and CIFAR-100, to verify real-time inference performance of our proposed CNN. Finally, we deploy these trained weights on NXP BlueBox which is an intelligent edge development platform designed for self-driving vehicles and UAVs, and conclusions will be extrapolated accordingly.

## 1. Introduction

Similar to natural intelligence demonstrated by living beings on planet Earth, Artificial Intelligence (AI) is the intellectuality demonstrated by machines involving consciousness and emotionality up to a very limited extent. Convolutional Neural Networks (CNN), a class of Deep Neural Networks (DNN), are a subset of Machine Learning (ML) which is a complex modus operandi for the realization of Artificial Intelligence (AI) by the machines. CNNs are becoming increasingly popular due to the advances in the field of embedded systems and edge devices, Internet of Things (IoT) and AI for performing fundamental Computer Vision (CV) tasks such as image classification, image segmentation and object detection for real-world applications such as autonomous cars, robots, and Unmanned Ariel Vehicles (UAV), commonly known as drones [1]. This rise in demand for AI and CV in addition to advancements in edge devices at local level such as the NXP BlueBox [2] which is an Automotive High-Performance Compute (AHPC) embedded platform designed specifically for autonomous (self-driving) vehicles by NXP Semiconductors N. V., has fueled the research and development of increasingly efficient CNNs for computationally resource constrained intelligent edge devices.

In this paper, we propose a new and an efficient version of CondenseNet CNN, dubbed EffCNet, designed for edge devices utilizing self-querying data augmentation and depthwise separable convolutional strategies to improve real-time inference performance as well as reduce the final trained model size, trainable parameters, and number of Floating-Point Operations (FLOPs). We will observe the results through training and validation statistics and then deploy these trained weights on NXP BlueBox 2.0 for real-time image classification purposes.

## 2. Representation Learning in Deep Neural Networks

Representation learning in deep neural networks is a modus operandi to allow a neural network model to automatically identify and learn different representations for feature detection and classification from the raw input data. It can be supervised i.e., features are learned through processing the labeled input data, or unsupervised i.e., low-dimensional features are discovered and learned through processing unlabeled input data.

Deep multilayer feed-forward neural networks are most commonly used to perform representation learning because relevant information is discovered, extracted and representations are learned in the hidden layer (s) and then used for classification in the output layer. Our proposed CNN, EffCNet, is based on the fundamentals of multilayer feed-forward neural networks for supervised representation learning.

The goal of representation learning in deep neural networks is to examine and learn hundreds of objects (classes) from thousands of images. However, this presents a significant challenge because these images contained a variety of noise and scale of different objects which overwhelm traditional classification schemes. Our work presented in this paper addresses this problem using a popular data augmentation scheme called AutoAugment [3] as explained in the next section.

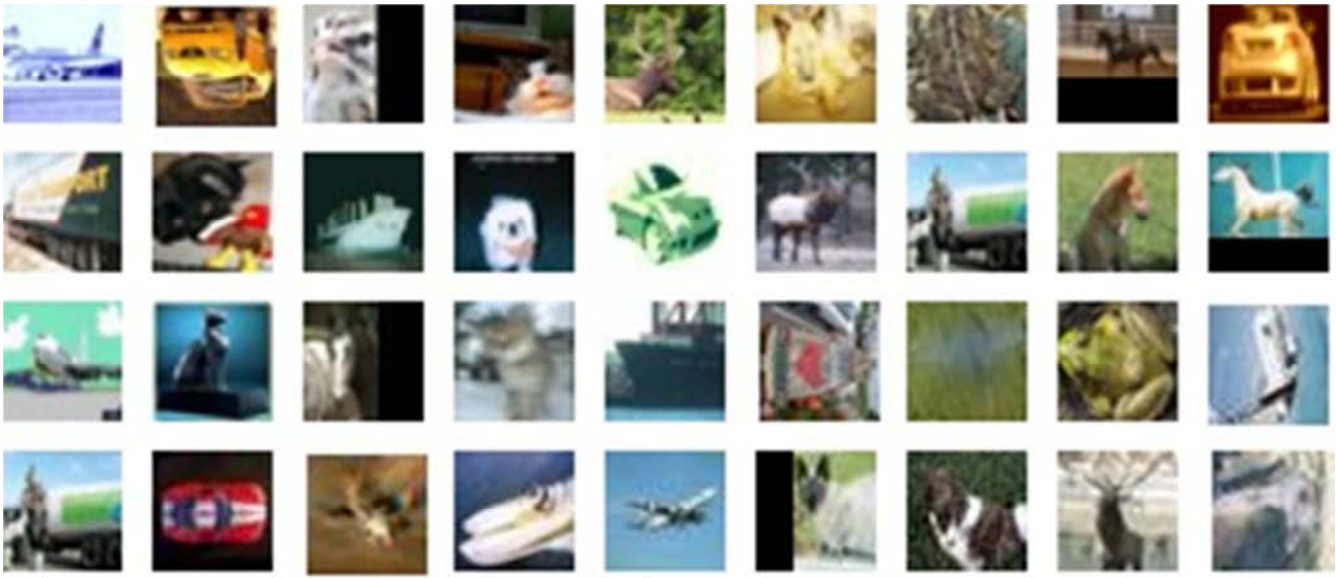

***Figure 1.*** *Color images of 32x32 pixels from different domains containing a variety of noise and irregularities.*

Deep learning neural network models for computer vision purposes have become increasingly successful and popular in the recent years due to advances in the algorithms utilized to exact features from the input data [4]. AlexNet, one of the world's most influential works in the field of AI, achieved a major milestone in 2012 when it was trained on ImageNet dataset and won the ImageNet LSVRC (Large Scale Visual Recognition Challenge) competition. It has, since then, inspired many works in the AI domain such as ZFNet [4] and VGG [5] neural networks which chose AlexNet as their baseline architecture which further resulted in improved performance and efficiency [6, 7]. However, at the time, very few neural network architectures and trained models offered the performance and efficiency so that they can be deployed for mass real-world commercial applications.

To address these issues, in 2016, an open-source machine learning library and a scientific computing framework called PyTorch was developed. PyTorch is based on the Torch library which is also an open-source machine learning library and a computing framework which was initially released in 2002. PyTorch is primarily developed by a team of researchers and engineers at the Facebook's AI Research (FAIR) lab and released under the license of Modified BSD [8]. The high-level features at the core of PyTorch are:

1. GPU accelerated tensor computing such as Numpy.
2. Tape-based autograd for automatic differentiation.

This has facilitated the use of PyTorch deep learning framework for commercial applications by big tech companies such as Tesla for their famous Autopilot system [9] for using in autonomous (self-driving) vehicles and Uber AI Lab for their open-source Pyro [10] which is a probabilistic programming language utilized for deep probabilistic modeling in order to make AI more accessible to the wider community. Our work presented within this paper is based on the latest stable release of the PyTorch framework at the time of writing this paper in order to introduce an improved and an efficient version of CondenseNet [11] CNN for edge devices utilizing self-querying data augmentation and depthwise separable convolutional strategies to improve real-time inference performance as well as reduce the final trained model size, trainable parameters, and FLOPs.

## 3. Background and Literature Review

In this section, we provide an overview of CondenseNet CNN architecture and its features as well as its pitfalls and scope of improvements to make it more computationally efficient.

### 3.1. Background

In 2016, ResNet [12] was introduced by G. Huang *et al.* which introduced identity short connections, a fundamental and groundbreaking technique to skip one or more layers in between. This facilitated the training of ResNet neural network on huge amounts of data more efficiently than ever before.

In 2017, the authors introduced DenseNet [13] by improving the technique of residual connections. In this architecture, authors proposed connecting each convolutional layer to every other layer in a feed-forward method so that features of the preceding layers are used as inputs to the current layer and the output of the current layer is used as input for all other subsequent layers. This resulted in reuse of tensors (neurons) in order to portray information at different depth levels.

### 3.2. Review of CondenseNet Architecture

In 2018, G. Huang *et al.* introduced CondenseNet, an improved version of DenseNet. It can be viewed as a form of pruning connections between several layers and blocks. In CondenseNet, classical convolutions are replaced with standard grouped convolutions ($G$). Furthermore, to avoid individual groups of channels being trained separately due to channel separation based on their offsets, authors propose learning of mapping between groups and channels by allowing layers from different groups to be connected directly as well. This technique is called learned grouped convolutions.

In CondenseNet, a new parameter called condensation factor ($C$) is introduced so that $G = C$. Furthermore, the growth rate increases exponentially so that a channel at $d$ depth has output channels: $2^d x_0$, where $x_0$ is any constant.

Each block in CondenseNet is comprised of a learned grouped convolution followed by a channel shuffling layer and finally a standard grouped convolution layer. Channels with the lowest $L_1$ norm are pruned from the network during the training stage. Figure 2 below provides a visual comparison between blocks of DenseNet and CondenseNet during the training stage.

An important point to notice is that CondenseNet utilizes standard grouped convolutions which perform 1 x 1 convolutions first. The convolutional layers do not traverse through the channel's spatial dimensions in the network, thus missing out details resulting in loss in efficiency. To address this issue, we propose the implementation of Depthwise Separable Convolution [14] which we shall see in the following sections in this paper.

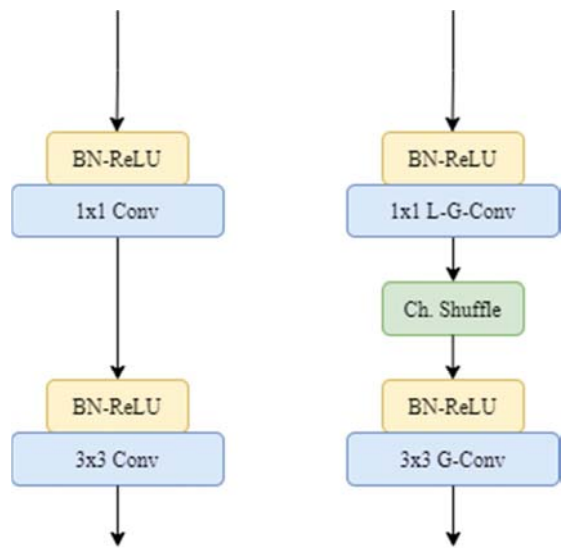

*Figure 2.* A visual representation of a dense block from DenseNet (left) in comparison to the dense block from CondenseNet (right) during the training stage.

## 4. EffCNet Architecture

In this section, we will propose a novel CNN architecture, named EffCNet, which stands for *Efficient CondenseNet* and has been inspired by the aforementioned prior works, and describe the architecture in detail.

### 4.1. Convolution Layers

The goal of EffCNet is to improve real-time inference performance as well as to reduce the final trained model size, trainable parameters, and Floating-Point Operations required to train the network from scratch. To achieve this goal, we first propose replacing the grouped convolutions with depthwise separable convolutions module. This module is made up of following two layers:

1. Depthwise convolutional layer which acts like a filtering layer. In standard convolutional layers such as the grouped convolution, the convolution operation is applied to all input channels whereas in case of depthwise convolution, a convolution is applied to an input channel independently.
2. Pointwise convolutional layer which acts like a combining layer. This layer combines all the outputs of depthwise separable convolution, resulting in a more efficient process of information extraction.

Figure 3 provides a visual representation of the depthwise separable convolution in 3D. This operation splits the kernel to be each used for depthwise (filtering) and pointwise (combining) convolutional layers. For example, in depthwise separable convolution, the channel 'Blue' in RGB channel is elongated to different shades of blue resulting in an efficient extraction of information at a reduced computational cost in comparison to the grouped convolutions, as observed through our experiment results in the following sections.

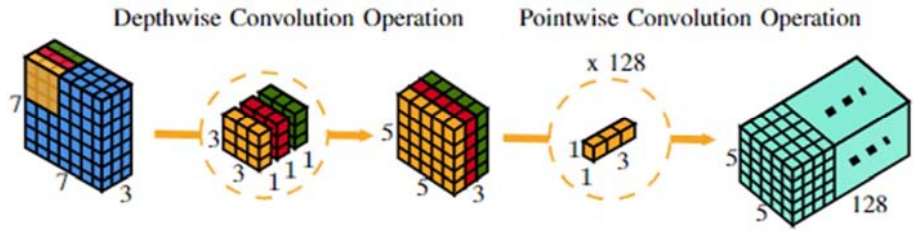

*Figure 3. A visual representation of a Depthwise Separable Convolution operation. This operation transforms the image only once and then this image is elongated up to 128 channels, as shown in the figure above.*

A standard convolutional filter $K$ can be mathematically represented as follows:

$$O_{k,l,n} = \sum_{i,j,m} K_{i,j,m,n} \cdot I_{k+i-1,l+j-1,m} \quad (1)$$

Here, the size of a standard convolutional filter is $S$ x $S$ x $X$ x $Y$, where $X$ is the number of input channels and $Y$ is the number of output channels with an input feature map denoted by $I$ of size $D_x$ x $D_x$ x $X$ that produces an output feature map denoted by $O$ of size $D_x$ x $D_x$ x $Y$.

A depthwise separable convolutional filter $K$ can be mathematically represented by factorizing equation (1) in to two parts for depthwise and pointwise operations as follows:

$$\hat{O}_{k,l,m} = \sum_{i,j} \hat{K}_{i,j,m} \cdot I_{k+i-1,l+j-1,m} \quad (2)$$

In this step, one 3 x 3 depthwise convolutional filter $\hat{K}$ is applied for each input channel and then, in the second step, a 1 x 1 pointwise convolutional filter $\widetilde{K}$ is applied to combine all outputs generated by equation (2). The second step can be mathematically represented by equation (3) as follows:

$$O_{k,l,n} = \sum_m \widetilde{K}_{m,n} \cdot \hat{O}_{k-1,l-1,m} \quad (3)$$

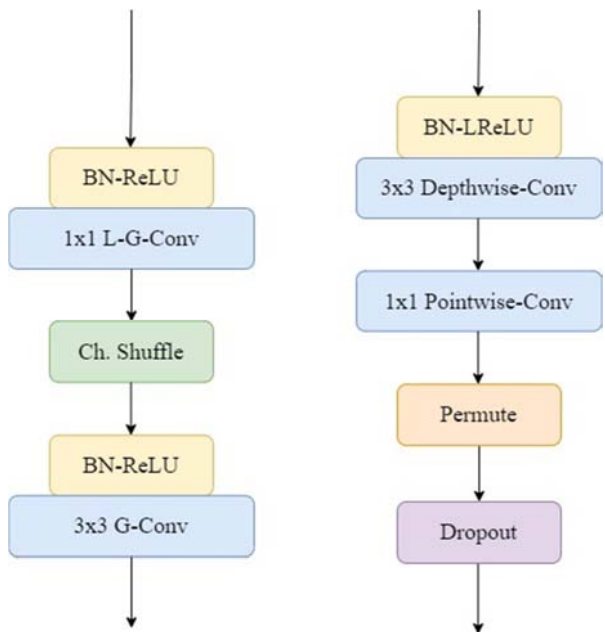

*Figure 4. A visual representation of a dense block from CondenseNet (left) in comparison to the dense block from our proposed CNN: EffCNet (right) during the training stage.*

This approach results in an efficient computation requiring fewer number of trainable parameters and FLOPs compared to the grouped convolution utilized in the baseline architecture of our CNN proposed within this paper.

Each block in EffCNet is comprised of a depthwise convolutional layer followed by a pointwise convolutional layer which is then followed by permute and dropout layers to prevent the model from overfitting. Figure 4 above provides a visual comparison between the blocks of CondenseNet and EffCNet during the training stage. Next, we replace and introduce Leaky-ReLU (LReLU) [15] non-linear activation into the design of our proposed CNN as discussed below.

### 4.2. Non-Linear Activation Function

For a particular given input or a set of given inputs, activation functions determine the output of that node in artificial neural networks. Non-linear activation functions facilitate the use of backpropagation algorithm for training multi-layer feed-forward deep neural networks. Non-linear activation functions have a derivative function relating to the inputs which facilitates error signal calculation by correlating the output of the network with the desired output. This error signal is then propagated backwards through the network and the weights are adjusted accordingly.

Rectified Linear Unit (ReLU) is a non-linear activation function [16] which is widely utilized by state-of-the-art neural networks because it is fast, simple, and empirical data suggest that DNNs with ReLU activation function converge quickly as opposed to using a DNN with sigmoid activation function. CondenseNet, the baseline architecture, utilizes ReLU activation function throughout its network. The major disadvantage of ReLU is famously known as the 'Dying ReLU' problem [17]. As the neurons become zero or negative, the corresponding function gradient becomes zero and the output remains the same i.e., these neurons no longer participate in discriminating inputs. Thus, rendering them dead with no recourse to recovery. The gradient descent learning no longer alters the weights and these neurons stop learning.

Our work presented within this paper utilizes Leaky Rectified Linear Activation (LReLU) function throughout the network to address the 'Dying ReLU' problem. Like ReLU, LReLU is also a non-linear activation function. However, unlike ReLU, LReLU activation function provides a small positive gradient for neurons with negative inputs so that they can recover instead of getting deactivated. LReLU can be mathematically represented as follows:

$$f(x) = \begin{cases} 0.01x, x < 0 \\ x, otherwise \end{cases} \quad (4)$$

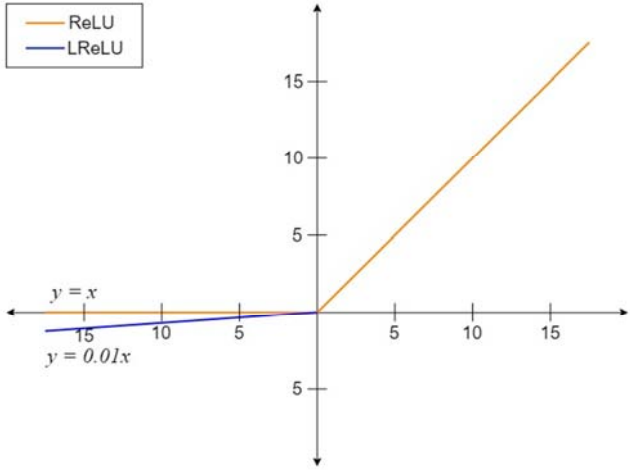

*Figure 5. Graphical representation demonstrating difference between ReLU and LReLU activation functions.*

Figure 5 below provides a visual representation of difference between ReLU and LReLU activation function in a graphical form.

Our experiments in the following sections demonstrate that this approach was able to achieve a small gain in Top-1 and Top-5 accuracies for both, CIFAR-10, and CIFAR-100 datasets.

### 4.3. Data Augmentation

Data augmentation is one of the many effective ways to improve accuracy of a neural network for image classification purposes. The target dataset is pre-processed to increase the diversity and amount of data by random augmentation [18].

Our work presented within this paper utilizes a well-known data augmentation strategy known as AutoAugment [3]. This strategy autonomously looks for advanced data augmentation policies which contain sub-policies. A sub-policy is arbitrarily chosen for each image in each mini batch. Each sub-policy is comprised of two distinct operations such that each operation on an image corresponds to the type of the operation and its probability and magnitude. Rotation, sheering, and translation are some of the different types of image processing operations within each sub-policy. A search algorithm is then utilized to determine a suitable policy for the neural network so that a best validation accuracy can be achieved on a chosen dataset.

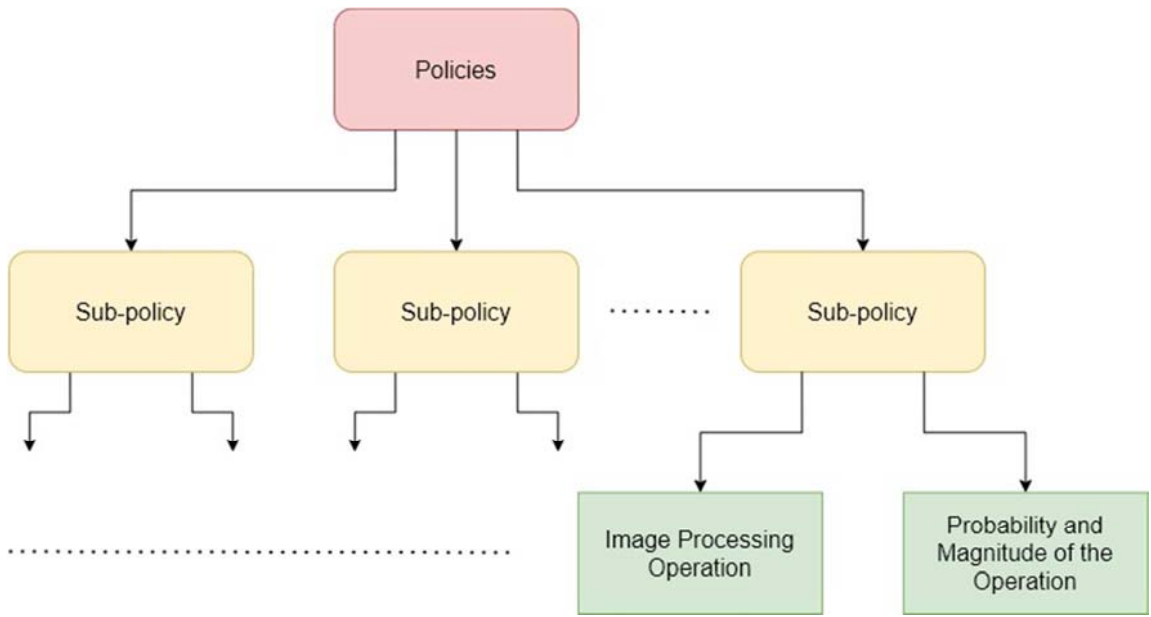

*Figure 6. Graphical representation auto augmentation strategy.*

## 5. Intelligent Edge Development Platform – NXP BlueBox 2.0

In this section, we provide a brief overview of the NXP BlueBox 2.0 hardware which we use to deploy the trained weights. NXP BlueBox is an intelligent edge development platform designed for self-driving vehicles and UAVs.

### 5.1. A Brief Overview

NXP BlueBox is an intelligent edge development and sensor-fusion platform designed and developed by NXP Semiconductors in 2016 for automotive applications such as self-driving (autonomous) vehicles. NXP BlueBox generation 1 was first introduced in 2016 at the NXP Technology Forum in Austin, Texas, USA [2]. In the following months, NXP introduced BlueBox generation 2 (also known as BlueBox 2.0) with three brand new ARM-based processors:
1. S32V234 for computer vision processing,
2. LS2084A for high performance computing, and
3. S32R274 for radar information processing.

NXP BlueBox 2.0 is an Automotive High-Performance Compute (AHPC) platform developed by adhering to strict automotive compliance and standards such as ASIL-B/C and ASIL-D for the reliability and safety of its operations in all conditions and workloads.

### 5.1.1. S32V234 (S32V) for Computer Vision Processing

The S32V234 processor was designed specifically for carrying out computer vision tasks. It houses a 4-core ARM Cortex A53 CPU operating at 1.0 GHz mated to another ARM Cortex M4 CPU to provide functional safety support along with a 4MB internal SRAM and a 32-bit LPDDR3 controller to support additional memory if needed. The S32V234 also offers an onboard Image Signal Processor (ISP) to meet/exceed ASIL-B/C automotive compliance and standards.

### 5.1.2. LS2084A (LS2) for High Performance Computing

The LS2084A processor was designed specifically to carry out high performance computational tasks. It houses an 8-core ARM Cortex A72 CPU operating at 1.8 GHz along with two 72 bytes DDR4 RAMs and supports LayerScape LX2 family, a class of high-performance multi-core processors developed by NXP. The LS2084A processor offers 15 years of reliability by meeting/exceeding Q100 Grade 3 automotive standards for reliability.

### 5.1.3. S32R274 (S32R) for Radar Information Processing

The S32R274 processor was designed specifically to perform real-time processor of radar information, such as LIDAR. It houses a dual-core Freescale PowerPC e200z4 CPU operating at 0.12 GHz mated to an additional checker core along with a 2MB flash memory and 1.5 MB of SRAM. The radar information processing is performed on-chip and meets/exceeds ASIL-D automotive compliance and standards.

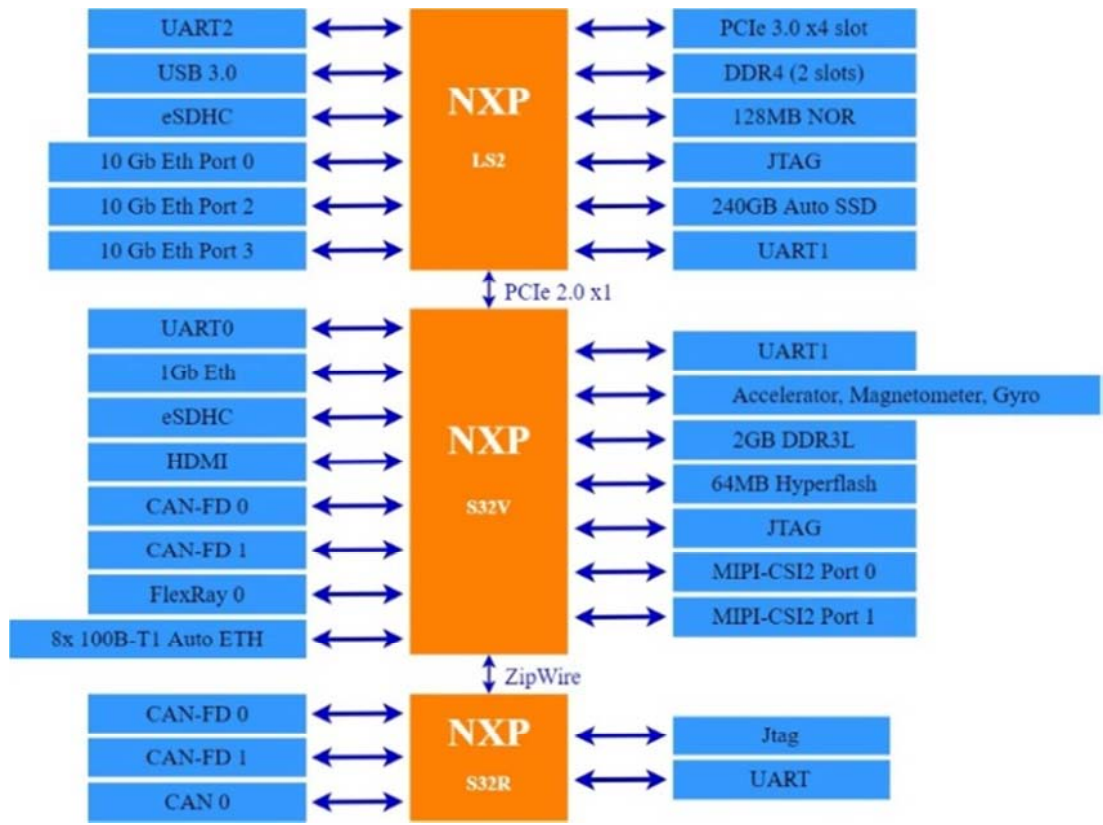

*Figure 7. High-level view of the NXP BlueBox 2.0 system architecture [2].*

### 5.2. RTMaps (Real-time Multi-sensor Applications)

Real-time Multi-sensor applications (RTMaps) is a remote studio GUI-based software designed and developed by Intempora to facilitate the development of automotive applications aimed at autonomous driving, sensor fusion and motion planning in addition to developing Advanced Driver-Assistance Systems (ADAS) applications [19].

RTMaps is a great tool to capture, view and process data obtained from multiple different sensors in in real-time. It is available for both, Windows and Linux operating systems and natively supports PyTorch and TensorFlow machine learning frameworks.

Our experiments presented within this paper in the following sections utilize open-source Python libraries and real-time image classification algorithm developed using Python scripting language. It is then deployed on to the NXP BlueBox 2.0 and conclusions are extrapolated accordingly.

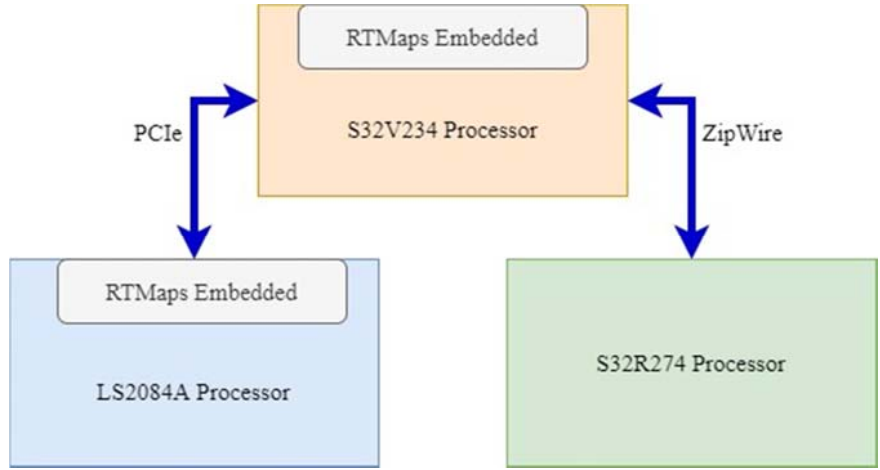

*Figure 8. A high-level view of RTMaps on NXP BlueBox 2.0.*

## 6. Cyberinfrastructure

In this section, we provide information on the hardware and software used for training and testing our CNN proposed.

### *6.1. Cyberinfrastructure for Training*

The following cyberinfrastructure for training is provided and managed by the Research Technologies division at the Indiana University which supported the work presented within this paper, in part by Shared University Research grants from IBM Inc. to Indiana University and Lilly Endowment Inc. through its support for the Indiana University Pervasive Technology Institute [20]:

1. Intel Xeon Gold 6126 12-core CPU, 128 GB RAM.
2. NVIDIA Tesla V100 GPU.
3. CUDA Toolkit 10.1.243.
4. PyTorch version 1.9.0.
5. Python version 3.7.9.

### *6.2. Cyberinfrastructure for Testing*

The following cyberinfrastructure for testing is provided and managed by the Internet of Things (IoT) Collaboratory at the Purdue School of Engineering and Technology at Indiana University Purdue University at Indianapolis, which also supported the work presented within this paper:

1. NXP BlueBox 2.0 development platform.
2. Intempora RTMaps Remote Studio version 4.8.0.
3. PyTorch version 1.9.0.
4. Python version 3.7.9.

## 7. Experiments and Results

In this section, we provide information and results based on extensive supervised image classification analyses conducted on two benchmarking datasets: CIFAR-10 and CIFAR-100, to verify real-time inference performance of our proposed CNN: EffCNet. EffCNet is designed and developed using the latest version of PyTorch framework at the time of writing of this paper and have been trained using the aforementioned cyberinfrastructure. We then deploy these trained weights on NXP BlueBox 2.0, and results are extrapolated accordingly as seen below.

### *7.1. CIFAR-10 Dataset*

CIFAR-10 dataset, introduced by A. Krizhevsky, is a labeled subset of the 80 million tiny images [21]. It contains 60,000 32x32 sized RGB images split in to 10 classes. These 60,000 images are further split in to 50,000 for training and 10,000 for testing the neural network.

EffCNet was trained for 200 epochs with a batch size of 64 and a single crop of input images. An image classification script in Python was then developed for this dataset using RTmaps remote studio software and then the corresponding trained weights were deployed on NXP BlueBox 2.0 autonomous vehicle development platform.

Table 1 provides an overview of the testing results in terms of number of trainable parameters, FLOPs, top-1 (the class with the highest probability matching the ground truth) and top-5 (first five classes with the highest probabilities matching the corresponding ground truth) accuracies. Figure 9, below, provides an RTMaps screenshot of the single image classification operation using EffCNet's trained weights on NXP BlueBox 2.0.

*Table 1. Comparison of Performance between CondenseNet (baseline architecture) and EffCNet (proposed CNN architecture).*

| Dataset | CNN Architecture | FLOPs (in millions) | Parameters (in millions) | Top-1 Accuracy | Top-5 Accuracy | Final Trained Model Size |
|---|---|---|---|---|---|---|
| CIFAR-10 | CondenseNet | 65.82 | 0.52 | 94.24% | 99.76% | 16.7 MB |
| | EffCNet | 61.01 | 0.46 | 94.15% | 99.83% | 2.1 MB |
| CIFAR-100 | CondenseNet | 65.85 | 0.55 | 75.06% | 93.17% | 17.5 MB |
| | EffCNet | 61.05 | 0.50 | 74.82% | 93.76% | 2.4 MB |

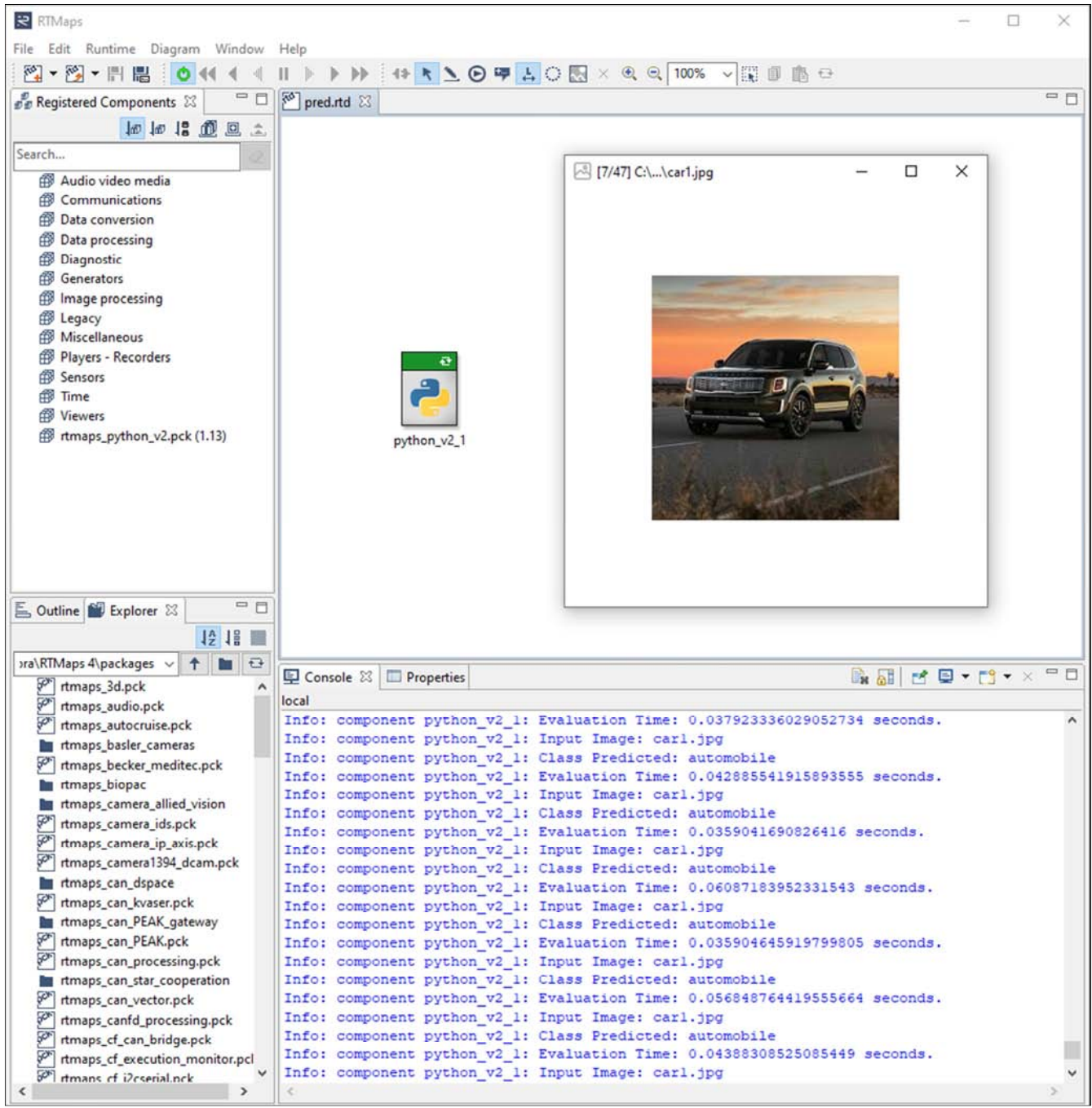

***Figure 9.*** *CIFAR-10 single image classification evaluation of EffCNet CNN when deployed using RTMaps on NXP BlueBox 2.0.*

### 7.2. CIFAR-100 Dataset

CIFAR-100 dataset, also introduced by A. Krizhevsky, is also a labeled subset of images from the 80 million tiny images collection [21]. It contains 60,000 32x32 sized RGB images split in to 100 classes. These 60,000 images are further split in to 50,000 for training and 10,000 for testing the neural network. Furthermore, classes and images of CIFAR-10 and CIFAR-100 datasets are mutually exclusive.

EffCNet was trained for 200 epochs with a batch size of 64 and a single crop of input images. An image classification script in Python was then developed for this dataset using RTmaps remote studio software and then the corresponding trained weights were deployed on NXP BlueBox 2.0 autonomous vehicle development platform.

Table 1, above, provides an overview of the testing results in terms of number of trainable parameters, FLOPs, top-1 (the class with the highest probability matching the ground truth), top-5 (first five classes with the highest probabilities matching the corresponding ground truth) accuracies and size of final trained model. Figure 10, below, provides an RTMaps screenshot of the single image classification operation using EffCNet's trained weights on NXP BlueBox 2.0.

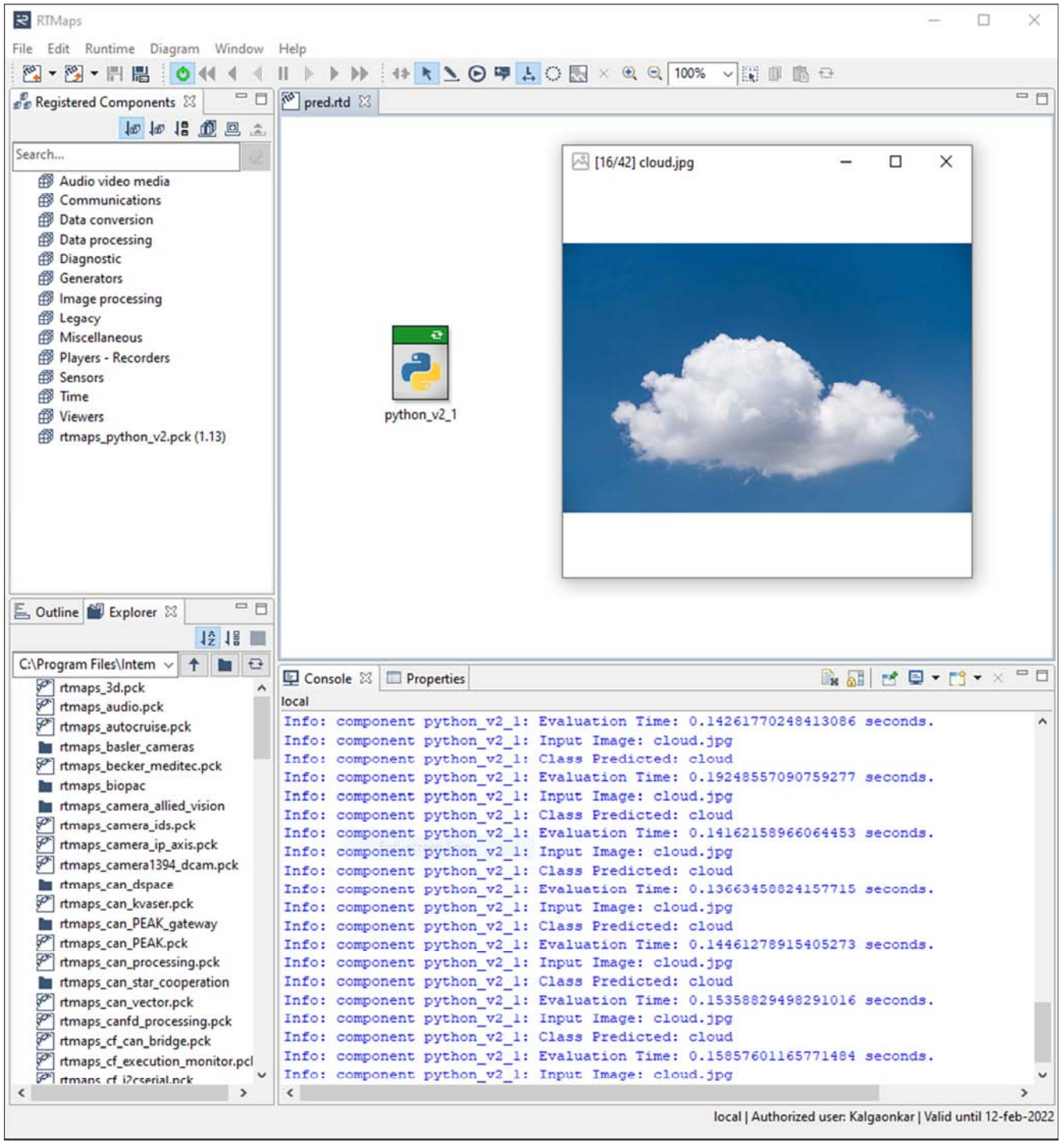

*Figure 10. CIFAR-100 single image classification evaluation of EffCNet CNN when deployed using RTMaps on NXP BlueBox 2.0.*

## 8. Conclusion

In this paper, we propose EffCNet CNN, a novel deep convolutional neural network architecture, which is an improved and an efficient version of CondenseNet CNN designed for edge devices utilizing self-querying data augmentation and depthwise separable convolutional strategies to improve real-time inference performance as well as reduce the final trained model size, trainable parameters, and Floating-Point Operations.

Extensive training from scratch and supervised image classification analyses were performed on the two benchmarking datasets: CIFAR-10 and CIFAR-100, to verify real-time single image classification performance of EffCNet. The performance results were then compared to the CondenseNet, the baseline architecture.

EffCNet achieves a top-1 accuracy of 94.15% on CIFAR-10 dataset and a top-5 accuracy of 74.82% on CIFAR-100 dataset. Our experiments on the NXP BlueBox 2.0 further demonstrate the efficiency of EffCNet's single image classification performance observed in terms of milliseconds. In the future, we shall explore image segmentation and object detection applications for EffCNet to further push the boundaries of the OpenCV realm.

## 9. Recommendations

Efficiency and accuracy play a critical role in developing a robust neural network algorithm for computer vision purposes since these algorithms play a crucial role in making life-critical decisions when deployed in real-world use cases such as autonomous (self-driving) cars, for example [9].

Our work and results presented within this paper provide some of the first experimental and implementation data of utilizing an automatic data augmentation scheme along with depthwise separable convolution technique.

Based on the data and results presented within this paper, following recommendations are given to the researchers depending on practical observations and analysis:
1) Perform experiments on other datasets such as YOLO (You Look Only Once) and Microsoft COCO (Common Objects in Context) for real-time object detection on embedded computing platforms such as the NXP BlueBox, Nvidia Jetson, etc.
2) Hyperparameter tuning is recommended to determine the best parameters required for training the neural network from scratch.
3) Implement EffCNet CNN algorithm for other OpenCV tasks such as image segmentation and object detection.

## Acknowledgements

The authors of this paper like to acknowledge the support of the IoT Collaboratory at IUPUI and the Indiana University Pervasive Technology Institute for providing supercomputing and storage resources that have contributed to all experiments and results reported within this paper.

## References

[1] Lotfi, A., Bouchachia, H., Gegov, A., Langensiepen, C., & McGinnity, M. (Eds.). (2019). Advances in Computational Intelligence Systems: Contributions Presented at the 18th UK Workshop on Computational Intelligence, September 5-7, 2018, Nottingham, UK. Springer International Publishing. https://doi.org/10.1007/978-3-319-97982-3.

[2] Cureton, C., & Douglas, M. (2019). Bluebox Deep Dive – NXP's AD Processing Platform. https://community.nxp.com/pwmxy87654/attachments/pwmxy87654/connects/258/1/AMF-AUT-T3652.pdf.

[3] Cubuk, E. D., Zoph, B., Mane, D., Vasudevan, V., & Le, Q. V. (2019). Auto Augment: Learning Augmentation Policies from Data. ArXiv: 1805.09501 [Cs, Stat]. http://arxiv.org/abs/1805.09501.

[4] Zeiler, M. D., & Fergus, R. (2014). Visualizing and Understanding Convolutional Networks. In D. Fleet, T. Pajdla, B. Schiele, & T. Tuytelaars (Eds.), Computer Vision – ECCV 2014 (pp. 818–833). Springer International Publishing. https://doi.org/10.1007/978-3-319-10590-1_53.

[5] Simonyan, K., & Zisserman, A. (2015). Very Deep Convolutional Networks for Large-Scale Image Recognition. ICLR.

[6] Deng, L. (2014). A tutorial survey of architectures, algorithms, and applications for deep learning. APSIPA Transactions on Signal and Information Processing, 3. https://doi.org/10.1017/atsip.2013.9.

[7] Krizhevsky, A., Sutskever, I., & Hinton, G. E. (2012). ImageNet Classification with Deep Convolutional Neural Networks. Advances in Neural Information Processing Systems, 25. https://papers.nips.cc/paper/2012/hash/c399862d3b9d6b76c8436e924a68c45b-Abstract.html.

[8] Paszke, A., Gross, S., Massa, F., Lerer, A., Bradbury, J., Chanan, G., Killeen, T., Lin, Z., Gimelshein, N., Antiga, L., Desmaison, A., Köpf, A., Yang, E., DeVito, Z., Raison, M., Tejani, A., Chilamkurthy, S., Steiner, B., Fang, L.,… Chintala, S. (2019). PyTorch: An Imperative Style, High-Performance Deep Learning Library. ArXiv: 1912.01703 [Cs, Stat]. http://arxiv.org/abs/1912.01703.

[9] Fridman, L., Ding, L., Jenik, B., & Reimer, B. (2019). Arguing Machines: Human Supervision of Black Box AI Systems That Make Life-Critical Decisions. 0–0. https://openaccess.thecvf.com/content_CVPRW_2019/html/WAD/Fridman_Arguing_Machines_Human_Supervision_of_Black_Box_AI_Systems_That_CVPRW_2019_paper.html.

[10] Bingham, E., Chen, J. P., Jankowiak, M., Obermeyer, F., Pradhan, N., Karaletsos, T., Singh, R., Szerlip, P., Horsfall, P., & Goodman, N. D. (2019). Pyro: Deep Universal Probabilistic Programming. The Journal of Machine Learning Research, 20.1 (2019), 973–978.

[11] Huang, G., Liu, S., Maaten, L. van der, & Weinberger, K. Q. (2018). CondenseNet: An Efficient DenseNet Using Learned Group Convolutions. 2018 IEEE/CVF Conference on Computer Vision and Pattern Recognition, 2752–2761. https://doi.org/10.1109/CVPR.2018.00291.

[12] He, K., Zhang, X., Ren, S., & Sun, J. (2016). Deep Residual Learning for Image Recognition. 2016 IEEE Conference on Computer Vision and Pattern Recognition (CVPR), 770–778. https://doi.org/10.1109/CVPR.2016.90.

[13] Huang, G., Liu, Z., Van Der Maaten, L., & Weinberger, K. Q. (2017). Densely Connected Convolutional Networks. 2017 IEEE Conference on Computer Vision and Pattern Recognition (CVPR), 2261–2269. https://doi.org/10.1109/CVPR.2017.243.

[14] Guo, Y., Li, Y., Feris, R., Wang, L., & Rosing, T. (2019). Depthwise Convolution is All You Need for Learning Multiple Visual Domains. ArXiv: 1902.00927 [Cs]. http://arxiv.org/abs/1902.00927.

[15] Xu, B., Wang, N., Chen, T., & Li, M. (2015). Empirical Evaluation of Rectified Activations in Convolutional Network. ArXiv: 1505.00853 [Cs, Stat]. http://arxiv.org/abs/1505.00853.

[16] Agarap, A. F. (2019). Deep Learning using Rectified Linear Units (ReLU). ArXiv: 1803.08375 [Cs, Stat]. http://arxiv.org/abs/1803.08375.

[17] Lu, L., Shin, Y., Su, Y., & Karniadakis, G. E. (2020). Dying ReLU and Initialization: Theory and Numerical Examples. Communications in Computational Physics, 28 (5), 1671–1706. https://doi.org/10.4208/cicp.OA-2020-0165.

[18] Simard, P. Y., Steinkraus, D., & Platt, J. C. (2003). Best practices for convolutional neural networks applied to visual document analysis. Seventh International Conference on Document Analysis and Recognition, 2003. Proceedings., 958–963. https://doi.org/10.1109/ICDAR.2003.1227801.

[19] El-Sharkawy, M. (2019, July). Speed is Key to Safety. DSPACE Magazine, 2/2019, 34–39.

[20] Stewart, C. A., Welch, V., Plale, B., Fox, G., Pierce, M., & Sterling, T. (2017). Indiana University Pervasive Technology Institute. https://doi.org/10.5967/K8G44NGB.

[21] Krizhevsky, A., & Hinton, G. (2010). Convolutional deep belief networks on cifar-10. Unpublished Manuscript, 40 (7), 1–9.